\documentclass[lettersize,journal]{IEEEtran}
\usepackage[colorlinks,urlcolor=blue,linkcolor=blue,citecolor=blue]{hyperref}
\expandafter\def\expandafter\UrlBreaks\expandafter{\UrlBreaks\do\/\do\*\do\-\do\~\do\'\do\"\do\-}
\usepackage{color}
\usepackage{booktabs}
\usepackage{tabularx}
\usepackage{multirow}
\usepackage{booktabs}
\usepackage{amsmath,amssymb,amsfonts}
\usepackage{latexsym}
\usepackage{subfig}
\usepackage{colortbl}
\usepackage{silence}
\usepackage{makecell}
\usepackage{balance}
\usepackage{graphicx}
\usepackage[table]{xcolor}

\newcolumntype{K}[1]{>{\centering\arraybackslash}p{#1}}
\newcolumntype{C}[1]{>{\centering\arraybackslash}m{#1}}
\newcolumntype{L}[1]{>{\raggedright\arraybackslash}m{#1}}
\definecolor{myblue}{RGB}{232, 240, 254}
\definecolor{myheader}{RGB}{91, 149, 249}

\hypersetup{nolinks=true}

\begin{document}

\title{Synthetic Image Verification \\ in the Era of Generative AI: \\
What Works and What Isn’t There Yet
}

\author{Diangarti Tariang$^1$, Riccardo Corvi$^1$, Davide Cozzolino$^1$, Giovanni Poggi$^1$, Koki Nagano$^2$, and Luisa Verdoliva$^1$
\IEEEmembership{$^1$University Federico II of Naples  \;\;\;  $^2$NVIDIA, USA}}


\maketitle

\begin{abstract}
In this work we present an overview of approaches for the detection and attribution of synthetic images and highlight their strengths and weaknesses. We also point out and discuss hot topics in this field and outline promising directions for future research.
\end{abstract}

\section{Introduction}
\begin{figure*}
    \centering
    \includegraphics[width=1\linewidth,trim=0 220 0 0, clip]{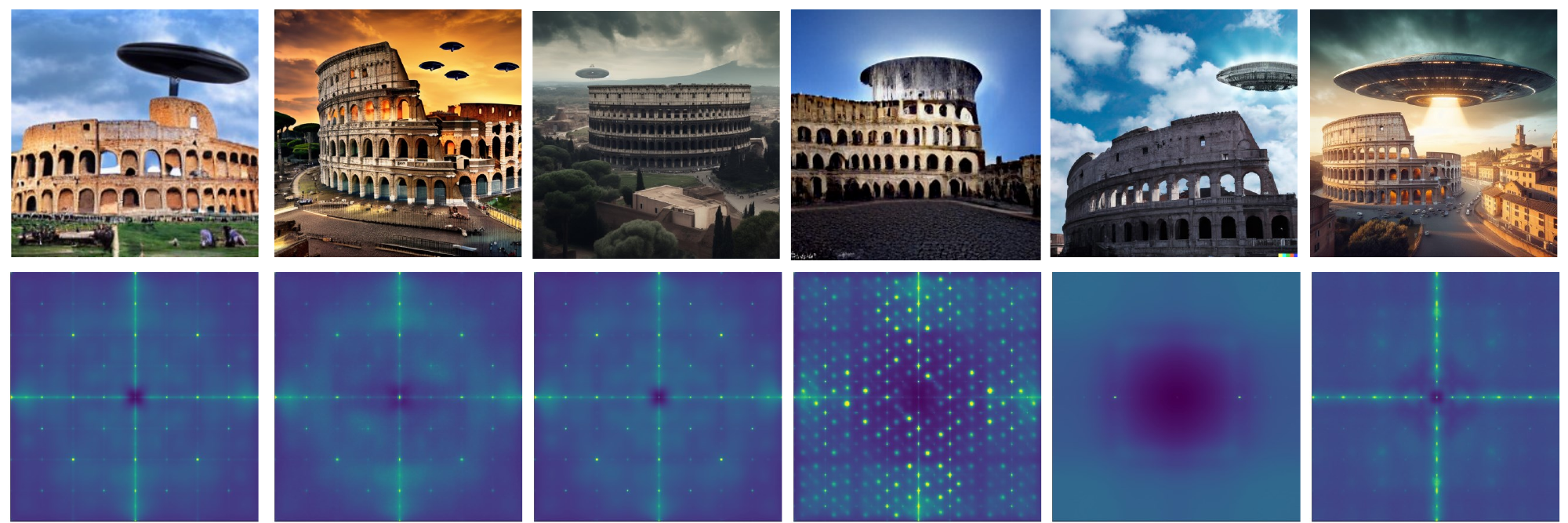}
    \caption{Top: examples of synthetic images, generated using (from left to right) Latent Diffusion, Stable Diffusion, Midjourney v5, DALL·E Mini, DALL·E 2, DALL·E 3. The prompt used for their generation is the following: {\em a photo of the Rome Colosseum with a UFO over it, detailed, 8k}. Bottom: Average Power Spectra of the artificial fingerprints for each of such model. Forensic artifacts are clearly visible as spectral peaks in the Fourier domain, stronger or weaker based on the specific model. We can observe that the first three images share very similar artifacts while the fingerprints of the three releases of DALL-E  differ greatly from one another, testifying to very different generative architectures\cite{corvi2023intriguing}.}
    \label{fig:os}
\end{figure*}

\IEEEPARstart{S}{ynthetic} media generation has seen tremendous progress in the span of just a few years. On the one hand, photorealism has fast improved with the advent of generative adversarial networks (GAN) and, more recently, diffusion models (DM).
On the other hand, the ease and flexibility of media generation has reached an unprecedented level.
Powered by large language models (LLMs), text-to-image synthesis tools allow the user to create from scratch and modify images at will by means of simple text instructions (see Fig.~\ref{fig:os}).
Generative AI offers numerous opportunities for many industries, from entertainment, to healthcare, to finance and manufacturing\cite{cardenuto2023age}. 
However, it can be used for all kinds of illicit purposes, especially to strengthen disinformation campaigns and political propaganda\cite{Epstein2023art}$^,$\cite{Barrett2023identifying}. Such goals can be now pursued faster than ever and on a large scale, with minimal human intervention and with results that are extremely realistic and well aligned with a specific narrative. This represents a serious threat to our society and justifies the growing focus on automated tools that distinguish synthetic images from natural ones. 

In this context, two slightly different objectives can be pursued:
{\it i) detection} provides a global score assessing the probability that the image under test is synthetic;
{\it ii) attribution} goes a step further and aims to trace the specific generative model used to synthesize the image.
By providing more specific information about the generation process, attribution validates the detection output and improves its interpretability.
Early generative AI approaches could introduce certain visual inconsistencies, such as asymmetries in shadows and reflected images. However, more recent ones can achieve an unprecedented level of realism, that make detection methods based on visual artifacts useless and push towards the discovery of invisible traces.
One possibility is to rely on subtle forensic traces left by the generation process.
In fact, each generative model leaves a sort of {\em artificial fingerprint}, which depends on the model architecture, the details of the synthesis process, and even on the training dataset.

In this article, we review the most effective approaches for synthetic image detection and attribution.
Then, we try to establish what can and cannot realistically be achieved with current methods and how reliable they are, especially when dealing with difficult real-world scenarios. 
Finally, we highlight the main current research challenges and indicate what we believe are the most interesting directions for future work. 

\section{Synthetic image generation}

Several powerful generative approaches have been proposed over the years, such as variational autoencoders, energy-based models, normalizing flows, generative adversarial networks, and diffusion models. Here we limit our attention to the last two approaches, both for their ability to generate high-quality images and for their easy support for text-image synthesis.
In fact, natural language-based image editing provides an unprecedented level of flexibility and control over the generation process, paving the way for new and more advanced applications.
Tab.~\ref{tab:ModelTables} lists the GAN- and DM-based image generators used in our experimental analysis.

\newcommand{\cm}[0]{\checkmark}
\begin{table}[t!]
    \centering
    \resizebox{1.0\linewidth}{!}{
    \setlength{\tabcolsep}{5pt}
    \renewcommand\arraystretch{0.0}
\begin{tabular}{lcccc}
       \toprule
  AI generative models & o & s & f & v  \\  \cmidrule(l){1-1} \cmidrule(l){2-5}
  \cite{karras2018progressive}      ProGAN           & \cm &     &     &     \\ 
  \cite{brock2018large}             BigGAN           & \cm &     &     &     \\
  \cite{choi2018stargan}            StarGAN          &     &     & \cm &     \\
  \cite{park2019semantic}           GauGAN           &     & \cm &     &     \\
  \cite{karras2020analyzing}        StyleGAN2        & \cm &     & \cm &     \\
  \cite{karras2021alias}            StyleGAN3        &     &     & \cm &     \\
  \cite{chan2022efficient}          EG3D             &     &     & \cm &     \\
  \cite{wang2023diffusion}          Diffusion-GAN    & \cm &     &     &     \\
  \cite{sauer2023stylegan}          StyleGAN-T       &     & \cm &     &     \\
  \cite{tao2023galip}               GALIP            &     & \cm &     &     \\ 
  \cite{kang2023scaling}            GigaGAN          & \cm &     &     & \cm \\  \cmidrule(l){1-1} \cmidrule(l){2-5}
  \cite{ho2020denoising}            DDPM             &     &     & \cm &     \\
  \cite{dhariwal2021diffusion}      ADM              & \cm &     &     &     \\
  \cite{song2021score}              Score-SDE        &     &     & \cm &     \\
  \cite{nichol2021glide}            GLIDE            &     & \cm &     &     \\
  \cite{rombach2022high}            Latent Diff.     & \cm & \cm & \cm &     \\
  \cite{ramesh2022hierarchical}     DALL·E 2         &     &     &     & \cm \\
  \cite{stablediffusion2}           Stable Diff.     &     &     &     & \cm \\
  \cite{balaji2022ediffi}           eDiff-I          &     & \cm &     &     \\ 
  \cite{peebles2022scalable}        DiT              & \cm &     &     &     \\
  \cite{brooks2023instructpix2pix}  InstructPix2Pix  &     &     &     & \cm \\
  \cite{deepfloydif}                DeepFloyd IF     &     &     &     & \cm \\
  \cite{podell2023sdxl}             SDXL             &     &     &     & \cm \\
  \cite{dalle3}                     DALL·E 3         &     &     &     & \cm \\
 \bottomrule
 \end{tabular}
}
\caption{List of the AI generative models analyzed in this work. Each model generates images with different content, as specified in the following: generic objects (o) by training on ImageNet/LSUN, faces (f) by training on FFHQ/CelebA, scenes (s) by training on COCO, various (v) includes objects, faces and scenes.}
    \label{tab:ModelTables}
\end{table}

\subsection{Generative Adversarial Networks}
GANs exploit the adversarial game between two networks, a generator that creates synthetic images and a discriminator that tries to distinguish them from natural images~\cite{goodfellow2014generative}.
The two networks are trained jointly with a min-max game: as the discriminator becomes more effective, so does the generator, creating increasingly realistic samples over time.
Among the first successful GAN-based methods, we mention BigGAN, a class-conditional image generator proposed by Brock et al.,
and ProGAN, proposed by Karras et al. in 2018, which produces high-quality images using a fast and stable training procedure that increases resolution over time.
Further improvements of the latter led to the StyleGAN family, where the convolutional kernels of the generator are controlled by latent code, allowing tight control of the synthesis process.
Appreciable results were also obtained in 3D synthesis using EG3D, a method capable of producing multi-view-consistent renderings and detailed geometry of a synthetic face.
Recently, the main research focus has shifted to text-based image synthesis, and several GAN-based methods have been proposed for this purpose.
Both StyleGAN-T and GALIP are jointly trained on images and text descriptions, using CLIP (Contrastive Language-Image Pre-Training)~\cite{radford2021learning} as the underlying language model, and produce high-quality samples in a controllable manner.
A further improvement has been made by GigaGAN, the first GAN-based method trained on billions of real-world images, which is capable of synthesizing high-resolution images very quickly, also supporting latent interpolation and stylization.

\subsection{Diffusion Models}
At the core of diffusion models are two interconnected stochastic processes.
A forward process transforms natural images into random noise by adding Gaussian noise in small steps. The samples generated in a single step of the forward process are then used to train a neural network that inverts that step, removing some noise from the input sample.
A chain of such networks performs the backward process, gradually converting input Gaussian noise into synthetic images.
The quality of images generated by diffusion models is comparable to that of GANs and better than that of other approaches.\cite{ho2020denoising,dhariwal2021diffusion}
Furthermore, training is easier and more stable than GANs, without mode collapse, although more time-consuming.
More importantly, with their flexibility, DMs provide ideal support for text-image synthesis, enabling the generation of complex images based on diverse and arbitrary text descriptions.
All this has revolutionized the way of tackling complex generative artificial intelligence tasks, and lead to many different architectures for text-image synthesis.

Most of these models rely on U-Net and its variations as a backbone, like GLIDE and DALL-E 2, that use a text encoder to condition generation on natural language descriptions based on CLIP.
The more recent Ediff-I adopts multiple U-Net models specialized for different synthesis stages.
To reduce computational costs, Latent DM combines a diffusion model with a variational autoencoder: the former operates in a low-dimensional space to generate the latent vector needed by the latter.
A noteworthy model of this class is Stable Diffusion, which is part of an open-source project and is trained on the 5.85 billion images of the LAION dataset \cite{schuhmann2021laion}.
A fast solution that can well generalize to real images and user-written instructions is Instructpix2pix, that combines the knowledge of GPT-3 and Stable Diffusion.
Stable Diffusion XL leverages a three times larger UNet backbone to generate very high resolution images. 
In Diffusion Transformers (DiTs), instead, the usual U-Net backbone is replaced by a transformer.
Inspired by Saharia et al., DeepFloyd IF has built a model, where the generation process includes a cascade of multiple networks \cite{Saharia2022photorealistic}.

\subsection{Forensic artifacts}
The images generated by synthetic architectures often featured visual artifacts such as unnatural colors or incorrect perspectives and shadows. 
These semantic inconsistencies are typically referred to as high-level artifacts.
Some models however produce visually perfect images without any obvious sign of their synthetic nature causing great concern among end users.
With the right methodology, it is still possible to tell apart real from synthetic images, even when these latter look perfectly realistic.
The general principle is that each image bears with it a number of distinctive marks related to the acquisition or generation process, which can be exploited to trace back its origin.
This is well-known for physical devices, where both hardware and software leave precious forensic traces. However, something very similar happens also for synthetic images.
AI-based generative models use complex processing chains involving a large number of specific processes, including filtering, pooling, downsampling and upsampling.
All such processes leave peculiar marks in the images which may be exploited to accomplish multiple forensic tasks, from source identification to forgery detection \cite{Chen2008determining}.
These are imperceptible traces, called also low-level artifacts, that can be exposed only by means of statistical analyses.
In the frequency domain, model-related artifact can be often spotted as strong peaks in the power spectra of noise residuals (see Fig.~\ref{fig:os}, bottom).
Furthermore, it has been clearly shown that synthetic generators struggle to perfectly reproduce  
the spectral distributions of the real data used for training in the medium/high frequencies \cite{corvi2023intriguing}$^,$\cite{durall2020watch}.

\begin{figure*}
    \centering
    \includegraphics[width=1\linewidth,trim=0 200 0 0,clip]{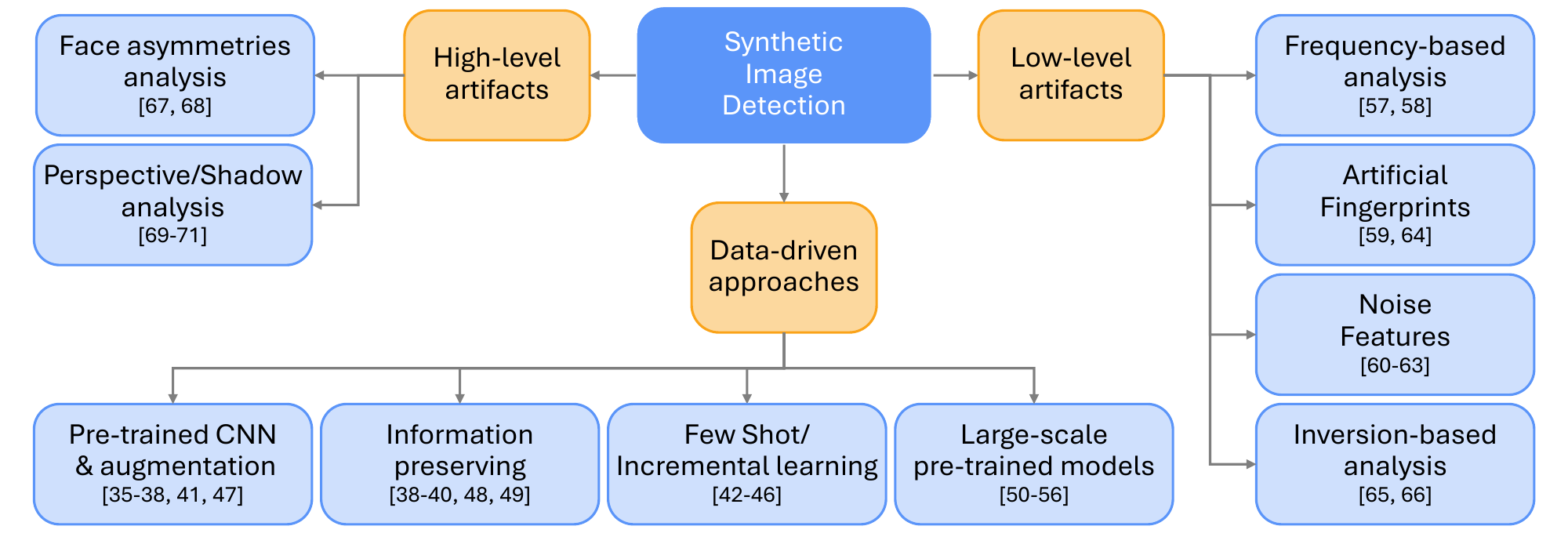}
    \caption{Taxonomy of synthetic image detection methods.}
    \label{fig:taxonomy}
\end{figure*}

\section{Synthetic image detection}

Early methods proposed to distinguish synthetic from real images relied on CNN-based architectures trained with large amounts of data. These methods work very well when test and training data are perfectly aligned but exhibit a significant performance drop in the presence of test-training mismatch. In particular, two major problems are lack of robustness and limited generalization ability. Robustness is necessary to withstand image impairments, like the re-compression and re-sizing of images posted on social networks, that weaken the subtle traces exploited by most classifiers. On the other hand, the ability to generalize allows the analysis of images that come from generators not seen during training.

In the following, we will review the main strategies proposed to handle such issues. It is worth underlining that most papers described in this Section focus on GANs, but in recent years there has been an ever increasing attention to DMs. However, methods conceived originally for GAN image detection usually turn out to work equally well on more recent AI-generative approaches.
A taxonomy of all the methods is presented in Figure~\ref{fig:taxonomy}.

\subsection{Data-driven methods}

A first strategy to achieve robustness to possible impairments is to leverage deep CNN architectures \cite{marra2018detection} and to include suitable augmentation during training.
In \cite{wang2020cnn} it was shown that good robustness can be achieved through simple augmentation with compressed and blurred images, even if the network is trained on a single generative architecture (ProGAN). A qualifying aspect of the proposal was also the training set diversity, ensured by the use of 20 different categories of images. Subsequent papers \cite{mandelli2022detecting,gragnaniello2021GAN} confirmed this to be a key factor to improve generalization ability. The adoption of large datasets for model pre-training appears to be important too. Extreme augmentation, instead, ensures only marginal gains in robustness but improves generalization to unseen models \cite{gragnaniello2021GAN}.

Another golden rule that applies equally well for GAN and DM image detection is to avoid any loss of information. This holds both during training and test, and regards all layers of the detector architecture, especially those closest to the input. 
In \cite{chai2020what} this is achieved by using a patch-based classifier and avoiding image resizing, in order not to erase the subtle traces left by the generation process.
To preserve the invisible forensics cues, in \cite{gragnaniello2021GAN} it is explicitly suggested to: 
{\em   i)} train the network on randomly cropped patches;
{\em  ii)} make the final decision on the whole image by means of some fusion strategy; 
{\em iii)} avoid any down-sampling in the first layers of the network.
In \cite{ju2022fusing} patch-based analysis is further enhanced by combining it with global spatial information extracted from the whole image.

The main goal of \cite{chandrasegaran2022discovering}, instead, is to single out transferable features that allow for the design of universal detectors.
In particular, color is shown to be a critical transferable forensic feature, and used in a suitable data augmentation scheme.
Another path towards improved generalization is the use of few-shot or incremental learning strategies, as done in \cite{cozzolino2018forensictransfer,marra2019incremental,du2020towards,jeon2020tgd}. Of course, these methods need the availability of some example images from the new architectures, data that may not be available in the most challenging scenarios. Along this direction a recent work investigated whether a detector was able to perform correct detection in a simulated online framework, where the detector is regularly re-trained by preserving the temporal order of the synthetic generator release date \cite{epstein2023online}. Results show that generalization is good to unseen models as long as the architecture of unseen generators is similar to that of old ones.

In \cite{corvi2023detection} an investigative analysis is carried out with several CNN-based methods for diffusion model detection. The results show that calibration is critical for detectors to work across different generators, and also that some form of fusion strategy could help. 
In \cite{dogoulis2023improving} it is shown that only higher-quality images should be included in the training dataset to ensure generalization across different categories. 
Explainable AI, instead, has been only lately explored, with a few works that leverage 
Gradient Class Activation Mapping to interpret the results \cite{bird2024cifake}.

Studies that consider backbone architectures different from CNNs, like transformers and vision-language models, are carried out in \cite{ojha2023towards}, \cite{ricker2022towards,amoroso2023parents,zhu2024genimage,cozzolino2023raising}. In particular, in \cite{ojha2023towards}, the generalization ability of a CLIP-ViT model pre-trained on internet-scale image-text pairs is empirically demonstrated. Classification is performed using the fixed feature space of the model, and a very good performance is obtained when trained on GANs and tested on DMs. Alternative strategies are pursued in \cite{sha2022fake,liu2023forgery}, where the proposed detector leverages also on the corresponding prompt during training.
Finally, note that in \cite{ricker2022towards,amoroso2023parents,zhu2024genimage} large datasets of synthetic images are made available to the community to foster research on synthetic image detection for generative AI.

\subsection{Methods exploiting forensic cues}

In this section we describe methods that more explicitely rely on some specific forensic traces both low-level and high-level cues.

\subsubsection{Low-level artifacts}
As already shown in Fig.~\ref{fig:os}, AI-generated images show clear traces of their origin in the Fourier domain. These artifacts are due to the up-sampling operations typical of the synthesis network. Even when such peaks are absent, synthetic images differ significantly from natural images at the medium-high frequencies.\cite{durall2020watch,corvi2023intriguing}
The artifacts in the Fourier domain were first exploited in \cite{zhang2019detecting}. The idea was to simulate such artifacts and then use them to train a spectrum based classifier. A similar idea was also pursued in \cite{frank2020leveraging}, where synthetic images were obtained trough an adversarial autoencoder and in \cite{jeong2022FingerprintNet} that proposes to simulate the fingerprints from real images by leveraging various types of generative models.

High-frequency traces can be also exposed by suppressing the scene content and extracting the noise residuals \cite{tan2023learning}. In this domain, real and synthetic images show different inter-pixel dependencies, which can be exploited for detection \cite{zhong2023rich,tan2023rethinking}. Noise patterns can be also learned as done in \cite{liu2022detecting}, \cite{sinitsa2023deep} or estimated during the inverse diffusion process \cite{zhang2023diffusion}.
Indeed, the inversion process can be very useful for detection as shown in \cite{wang2023dire}. The idea is that, unlike real images, DM-generated images can be accurately reconstructed by a DM. Therefore, by measuring the error between an input image and its reconstruction counterpart it is possible to carry out detection. 

\subsubsection{High-level artifacts}
Some methods look for semantic errors, such as asymmetries in faces, wrong perspective, odd shadows. These works are either focused on faces \cite{matern2019exploiting,bohacek2023geometric} or on generic scenes \cite{farid2022lighting,farid2022perspective,sarkar2023shadows}.
In \cite{sarkar2023shadows}, based on the observation that shadow and perspective errors are systematic in diffusion models, a classifier is proposed that looks at the perspective field, at lines, and at the relations between detected objects and shadows.

\begin{figure}[t!]
    \centering
    \includegraphics[width=1.0\linewidth,trim=0 0 0 0, clip,page=1]{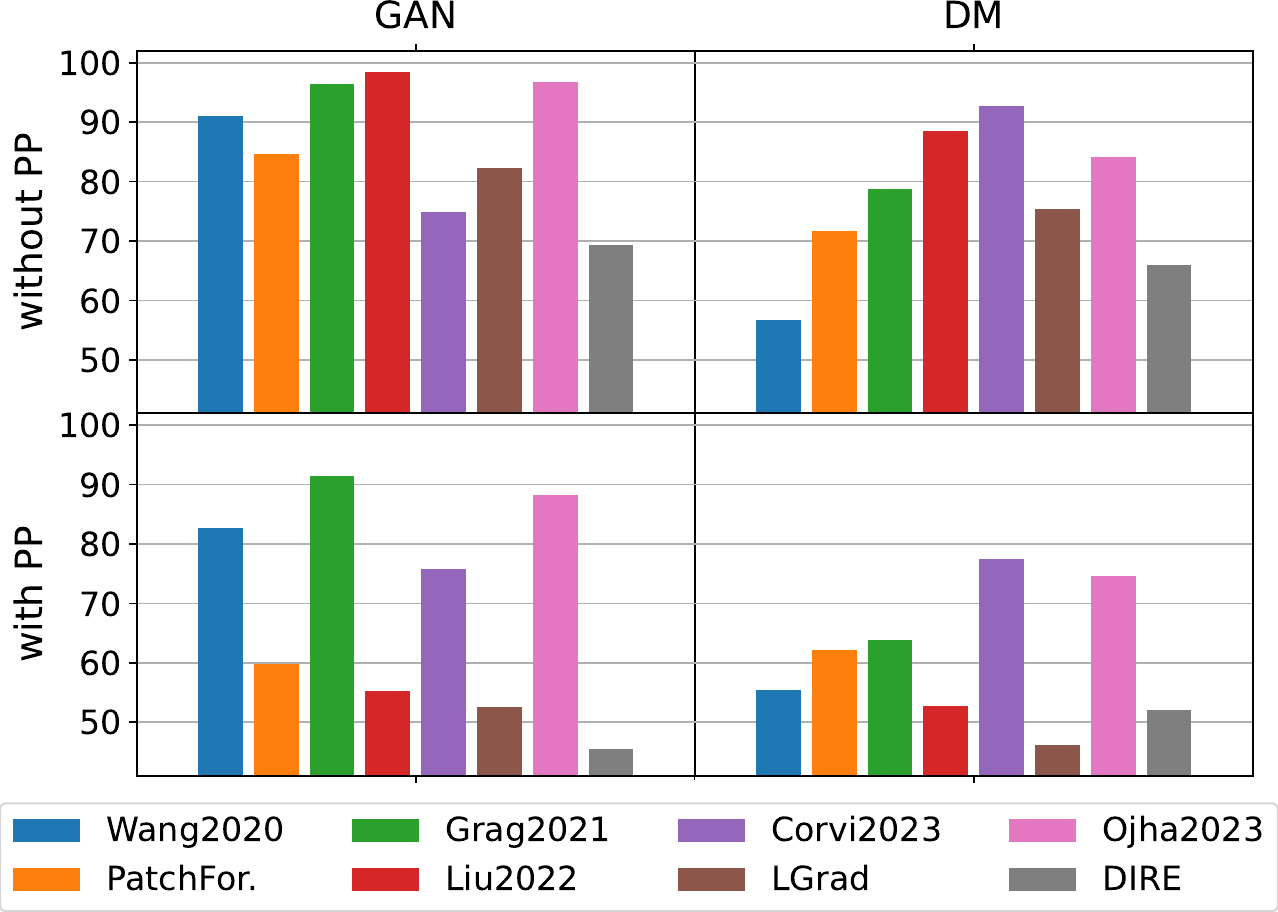} \\
    \caption{Synthetic image detection results in terms of AUC with and without post-processing (PP).}
    \label{fig:res}
\end{figure}

\subsection{Experimental evaluation}
In this Section we carry out a comparison of the methods proposed in the literature for which code and model were publicly available:
Wang2020 \cite{wang2020cnn},  PatchFor. \cite{chai2020what}, Grag2021 \cite{gragnaniello2021GAN}, Liu2022 \cite{liu2022detecting}, Corvi2023 \cite{corvi2023detection}, LGrad \cite{tan2023learning},  Ojha2023 \cite{ojha2023towards}, and DIRE \cite{wang2023dire}.
For each method we use the model already trained by the authors as proposed in their original paper.

\subsubsection{Generalization analysis}
In this Section, we show some experiments on generalization.
The test set comprises 1,000 synthetic images for each of the generators listed in Table~\ref{tab:ModelTables}.
To evaluate performance, the synthetic images of each generator are compared with real images with the same content, so as to avoid biases induced by different contents.
In particular, we used 5,000 real images, 1,000 per each dataset: LSUN, FFHQ, ImageNet, COCO and LAION.
For simplicity we aggregate results for all GAN-based generated images and all DM ones.

\begin{figure}[t!]
    \centering
     \includegraphics[width=1.0\linewidth,trim=0 0 0 0, clip, page=1]{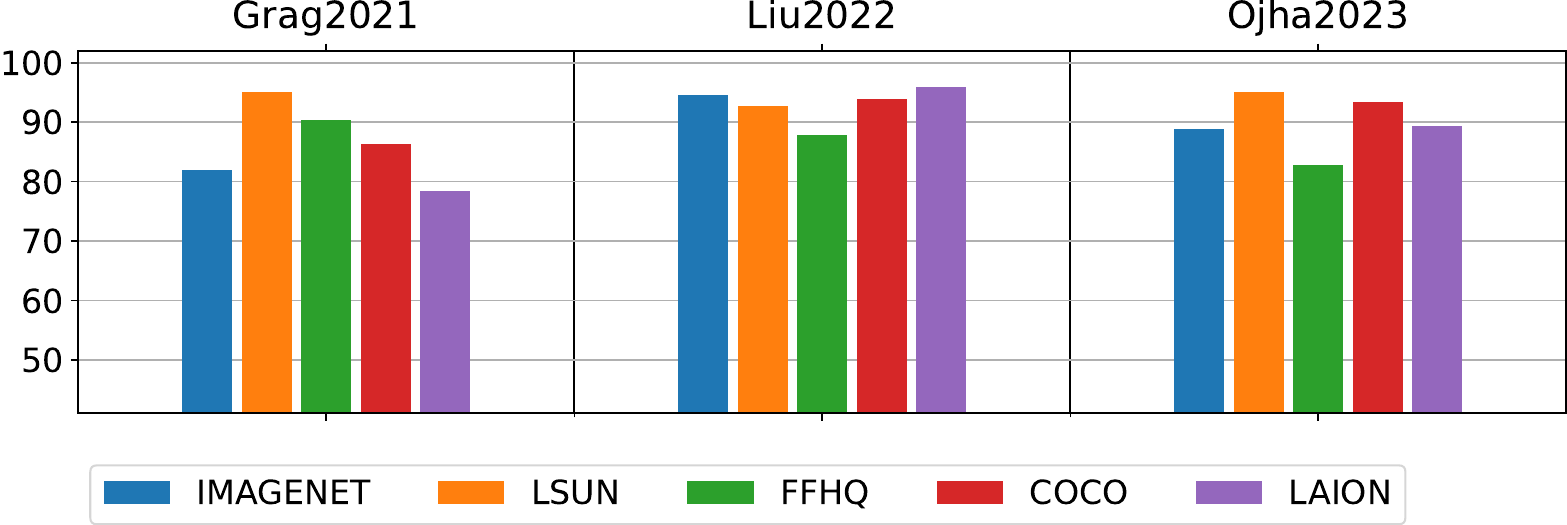} 
    \caption{Synthetic image detection results in terms of AUC for three methods over various contents.}
    \label{fig:resdb}
\end{figure}

To simulate a realistic scenario on the web, we also consider a post-processed (PP) version of the datasets. First, it is taken an image crop, that can vary in a range that goes from $\frac{5}{8}$ to the full image size. The image is then resized to $200\times200$ pixels and, finally, it is JPEG compressed with a random quality factor between $65$ and $100$.
Results are shown in terms of Area Under the ROC Curve (AUC) in Fig.~\ref{fig:res}. We can observe that DM images are harder to detect than GAN images. This can be explained by the fact that most detectors are trained on the ProGAN dataset except for Corvi2023 that is trained on Latent Diffusion. Overall the best generalization is ensured by Ojha2023, that relies on CLIP as backbone. However all performance figures worsen in the presence of a significant training-test mismatch (Fig.~\ref{fig:res}, bottom) and some detectors exhibit an AUC very close to 50\%, especially with DM images. Interestingly the performance does not vary across different categories as can be seen in Fig.~\ref{fig:resdb}. For example, both Ojha2023 and Liu2022 show very good performance on scenes or faces even though they are not trained for these categories.

\subsubsection{In the wild}
\begin{figure}[t!]
    \centering
     \includegraphics[width=1.0\linewidth,trim=0 0 0 0, clip, page=1]{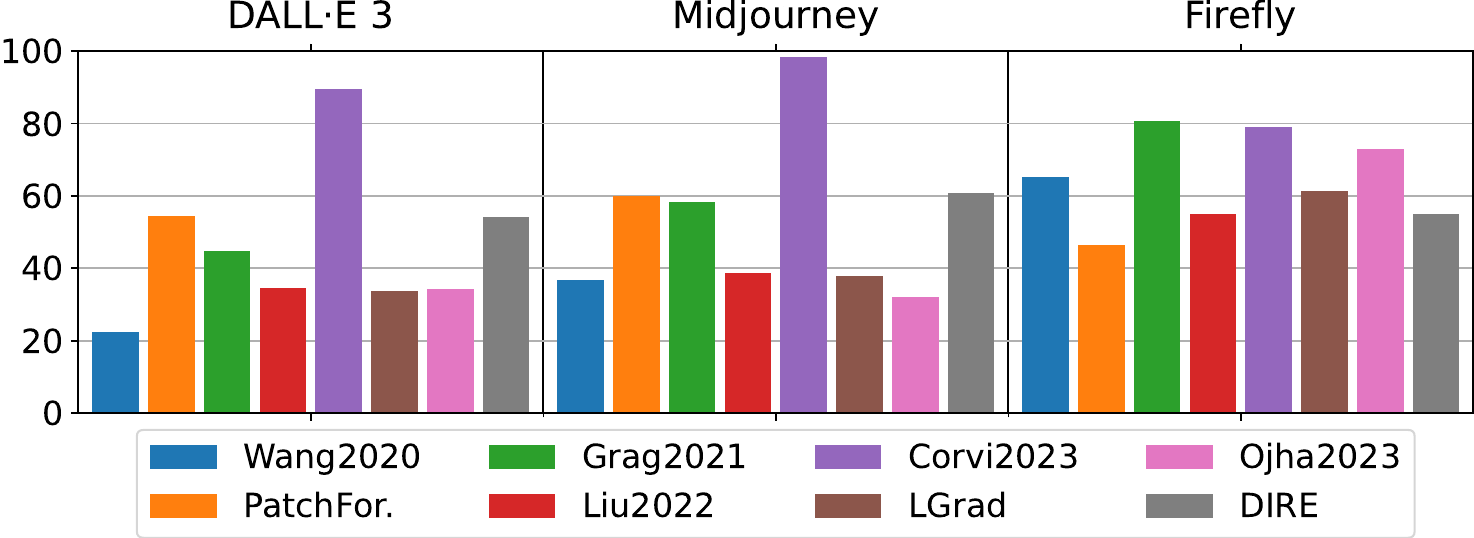} 
    \caption{Synthetic image detection results in terms of AUC on images from X generated by DALL·E 3, Midjourney and Firefly.}
    \label{fig:resother}
\end{figure}

To evaluate the performance in a more challenging situation, we downloaded a total of 2,000 images from a well known social network (X), both real and generated from Midjourney v5, DALL·E 3 and Firefly (according to their associated hashtag). Experimental results are shown in Fig.~\ref{fig:resother} in terms of AUC.
In this case, methods behave quite differently than before, with CNN-based architectures performing much better than the others. In particular, Corvi2023 has a near-perfect performance for Midjourney v5 and DALL·E 3. This is easy to explain if we observe again Fig.~\ref{fig:os}: the forensic artifacts for Midjourney v5 are almost identical to those embedded in Latent DM images, used in training.

\subsubsection{Calibration}
\begin{figure}[t!]
    \centering
     \includegraphics[width=1.0\linewidth,trim=0 0 0 0, clip, page=1]{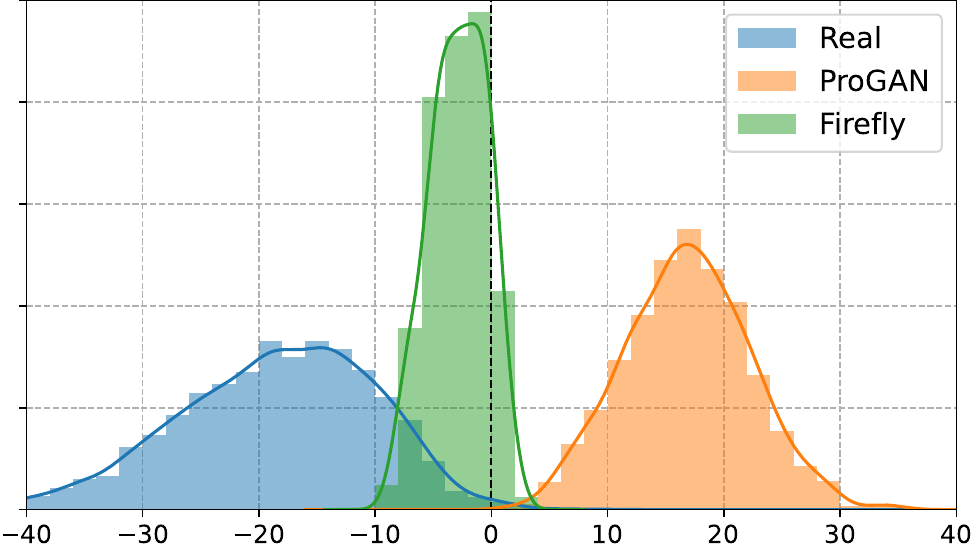} 
    \caption{Scores distribution provided by the method \cite{gragnaniello2021GAN} (trained on ProGAN) for real images, and for synthetic images from ProGAN and Firefly.}
    \label{fig:hist}
\end{figure}

We presented all the results in terms of AUC. However, when working in a practical scenario, it is necessary to set a threshold to discriminate between real and fake images. This is by no means a trivial problem: a fixed threshold is hardly appropriate in all  situations, especially if training and test data are misaligned, and could lead to disappointing results. An example is shown in Fig.~\ref{fig:hist}, where we show the distribution of the scores provided by Grag2021 trained on ProGAN. We can see that the distribution of ProGAN test images is very well separated from the real one if we choose a zero threshold. However, if we aim at distinguishing Firefly (not included in training), then a lower threshold should be used.

\section{Synthetic image attribution}

Image attribution extracts information about the provenance of the image by linking it to its generative model. If the "real" class is among those considered in the process, then the attribution includes automatically the detection.

\subsection{Artificial fingerprints}
Early research in the field of synthetic image attribution focused primarily on adapting successful techniques and methods from conventional multimedia forensics to this emerging domain. In particular, device and model fingerprints have proven to be extremely valuable for a wide range of forensic tasks and widely used especially for source identification.
The concept of device fingerprint was first introduced by Chen et al. \cite{Chen2008determining} who demonstrated that imperfections inherent in the camera sensor generated a unique and stable pattern in each captured image, a fingerprint in every way, called photo-response non-uniformity (PRNU) pattern. As already mentioned, similar fingerprints can also be extracted from synthetic images and represent a valuable tool for image attribution. The procedure requires removing the semantic content of the scene through a denoiser and then averaging the residual images. As the number of averaged images grows, a weak but stable quasi-periodic pattern can be observed \cite{marra2019DoGAN}.

Such fingerprints allow discriminating different models but can also provide information on the different datasets used for training. In fact, the same architecture, trained in different conditions, gives rise to slightly different fingerprints that allow fine-grained model authentication. Overall, the observed fingerprint turns out to depend on the specific architecture, the training dataset and also the training hyperparameters (e.g. learning rate, optimizer, training iterations) \cite{yu2019attributing}. 
A learned fingerprint is also extracted in \cite{ding2021does} through a hierarchical Bayesian approach. It does not depend on the initial seed, but is shown to change based on the training dataset and is robust to benign image transformations, such as compression, blurring and additive noise. The influence of the dataset was further analyzed highlighting possible biases in the generated data and extending the analysis of such traces to diffusion models\cite{corvi2023intriguing}. A slightly different perspective is taken in \cite{asnani2021reverse}, where a fingerprint estimation network is proposed to capture the unique patterns used to predict the network architecture and loss function that characterize a generative model. 
It is worth underlining that, in any case, these fingerprints are due to subtle traces present in the image, and therefore lay in the medium/high frequency bands. Therefore, while exhibiting a certain robustness, they can be vulnerable to adversarial attacks or can even be replaced by alien fingerprints extracted from real cameras \cite{Cozzolino2021SpoC}.

\subsection{Attribution via inversion}
Attribution can also be pursued by image inversion, since a generative model is not able to perfectly synthesize an image coming from another model nor it can perfectly reproduce a real image \cite{karras2019style}. The idea is to know a set of possible generators, both architectures and weights, and then compute the input data of the generator that allows to produce an image as similar as possible to the image under test. The likely source is the generator that ensures the minimum reconstruction error \cite{albright2019source}. A robustness analysis and the extension to an open set scenario can be respectively found in \cite{zhang2021attribution} and \cite{hirofumi2022did}.

\subsection{Attribution as a classification problem}
The source attribution problem can be simply considered as an $N$-ary classification, where the $N$ classes can be associated for example with different architectures. In \cite{bui2022repmix}, in order to learn an attribution model that is robust to different categories and possible perturbations, it has proposed a mix-up representation training strategy at the feature level. Interestingly, the approach can effectively handle both detection and attribution through a compound loss that takes into account the hierarchical nature of the problem. The image can be classified as real or fake and only in the latter case should the problem of attribution be considered. 
Instead in \cite{yang2022deepfake} a patchwise contrastive learning strategy is pursued with pre-training on image transformation classification that have been found to be similar to the generator components.

\subsection{Open-set methods}
When dealing with synthetic image attribution, several methods assume to work in a closed-set scenario, where test images were generated by a limited set of architectures, known in advance, whose samples were present in the training set. However, this scenario is far from realistic, given that new generation methods are proposed every day. In fact, images of new architectures may not even be available. In this condition, it is very important to be able to identify an out-of-distribution sample, recognizing that it does not belong to any of the classes observed in the training process. This more realistic scenario is known as open set recognition.

\begin{figure}[t!]
    \centering
    \small
    \setlength{\tabcolsep}{1pt}
    \begin{tabular}{c}
    Closed-set Accuracy \\
         \includegraphics[width=0.95\linewidth,trim=0 0 0 0, clip,page=1]{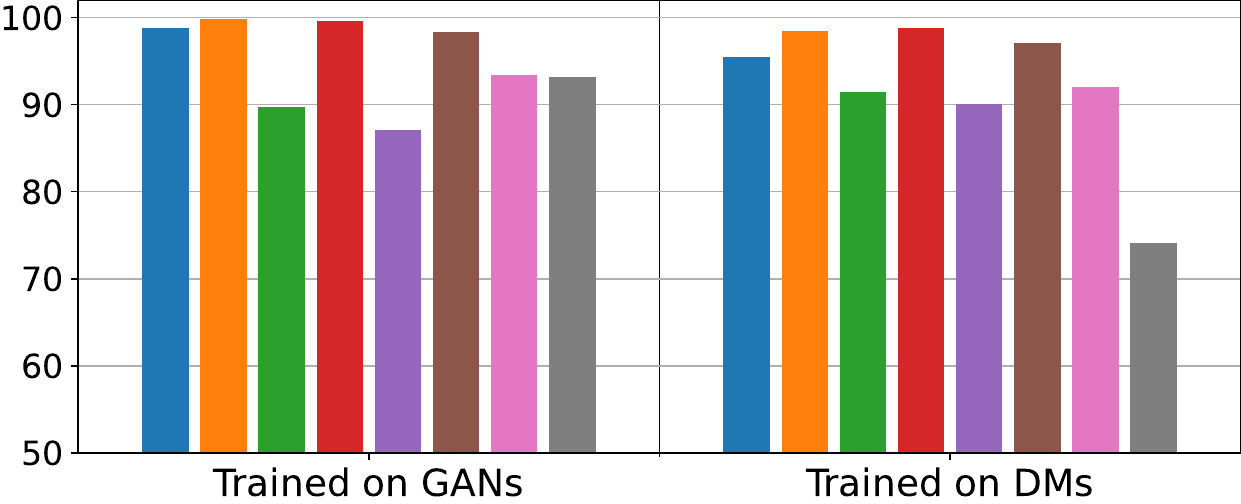}  \\
         Open-set AUC  \\
         \includegraphics[width=0.95\linewidth,trim=0 0 0 0, clip,page=2]{figures/results_att_bars.pdf}  \\[2mm]
         \hspace{0.5cm} \includegraphics[width=0.9\linewidth,trim=0 0 0 0, clip,page=4]{figures/results_att_bars.pdf}
    \end{tabular}
    \caption{synthetic image attribution results in both closed-set and open-set scenarios, considering two separate training sets (GANs an DMs).}
    \label{fig:att_result}
\end{figure}

A first open-set approach was proposed in \cite{girish2021towards} with an iterative algorithm that discovers new classes and re-trains the network using them as pseudo-labels. This method uses a fixed set of labeled images and then performs attribution on a set of unlabeled data through a clustering procedure. More recent works rely on classification with a rejection option, as in \cite{wang2023open} based on a hybrid ViT+ResNet50 architecture, or in \cite{fang2023open} where a metric-learning based embedding is developed to measure the similarity between the source generators of synthetic images.
Finally, in \cite{yang2023progressive} a progressive open space solution is proposed. The idea is to simulate unknown classes trough a set of augmentation models, based on lightweight networks that can model the traces of a variety of unknown models.

\subsection{Closed-set vs Open-set analysis}

In this Section we carry out a comparison of some of the methods proposed in the literature for which code was publicly available on-line: RepMix~\cite{bui2022repmix}, DNA-Det~\cite{yang2022deepfake}, GAN Discovery~\cite{girish2021towards}, Wang2023~\cite{wang2023open}, POSE~\cite{yang2023progressive}. All methods were re-trained using exactly the same data. Along with the methods specifically proposed for attribution, we also consider some of the best detection methods analyzed in the previous section re-trained in a multi-class configuration \cite{wang2020cnn,gragnaniello2021GAN,ojha2023towards}. 
We consider both closed-set and open-set scenarios and carry out attribution at architecture-level.
We evaluate the performance of closed-set attribution in terms of accuracy, a widely utilized metric for multi-classification problems.
For the open-set scenario, we measure the capacity to distinguish between in-distribution and out-of-distribution samples in terms of AUC, as is typically done in the literature \cite{wang2023open,fang2023open,yang2023progressive}.
More specifically, We analyze different situations as described below.

\begin{figure}[t!]
    \centering
    \setlength{\tabcolsep}{1pt}
    \begin{tabular}{cc}
    Closed-set Accuracy  & Open-set AUC   \\
         \includegraphics[width=0.49\linewidth,trim=0 0 0 0, clip,page=1]{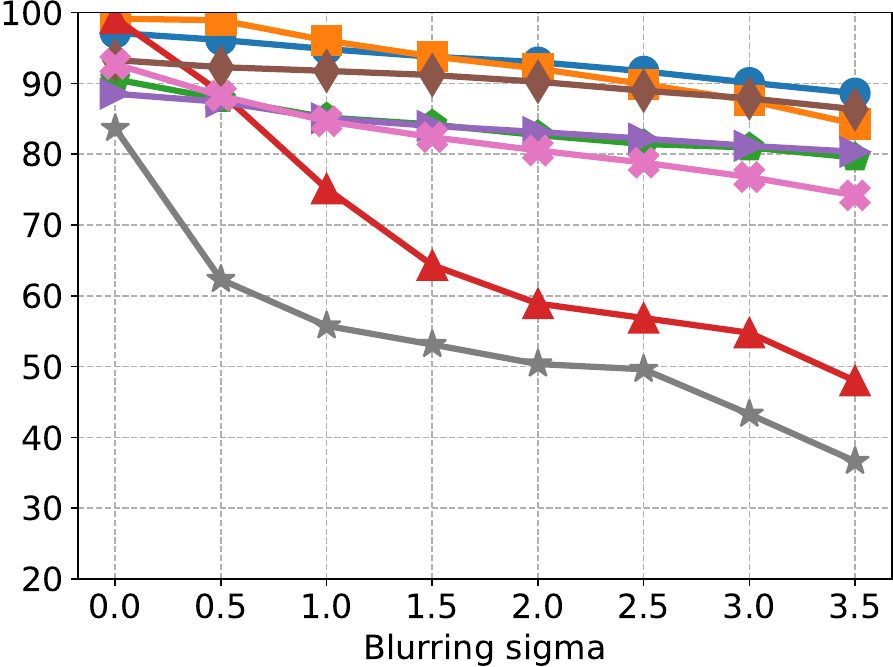} &
         \includegraphics[width=0.49\linewidth,trim=0 0 0 0, clip,page=2]{figures/results_att_blur.pdf}  \\
         \includegraphics[width=0.49\linewidth,trim=0 0 0 0, clip,page=1]{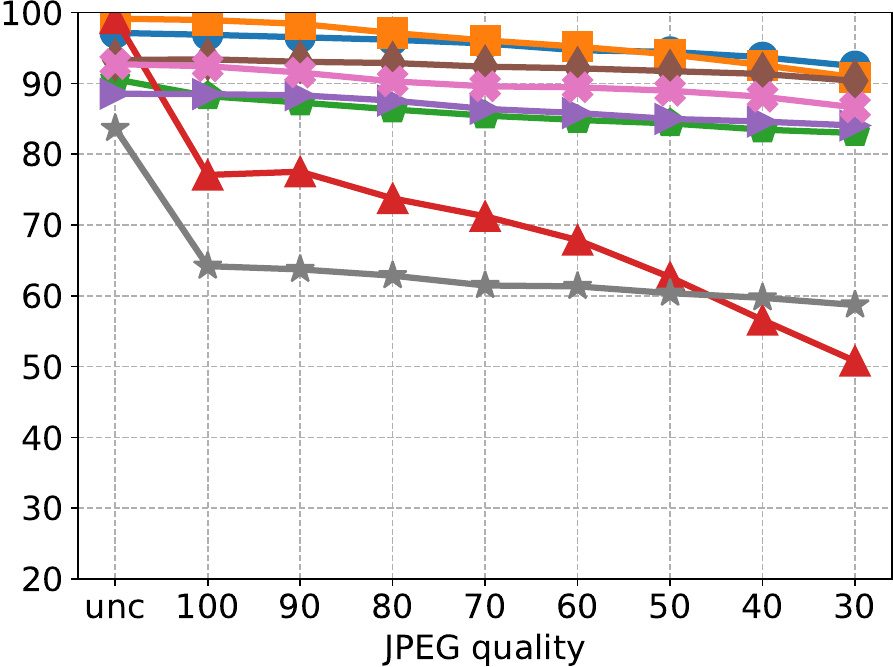} &
         \includegraphics[width=0.49\linewidth,trim=0 0 0 0, clip,page=2]{figures/results_att_jpeg.pdf} \\
         \multicolumn{2}{c}{\includegraphics[width=0.9\linewidth,trim=0 0 0 0, clip,page=4]{figures/results_att_jpeg.pdf}}
    \end{tabular}
    \caption{Robustness analysis in both closed-set and open-set scenarios, by varying Gaussian blurring (top) and JPEG compression quality (bottom).}
    \label{fig:att_robust}
\end{figure}

\subsubsection{GAN-based vs DM-based}
To analyze the possible different performances among GANs an DMs, we consider two separate sets of 7 synthetic generators, one is GAN-based (ProGAN, BigGAN, StarGAN, GauGAN, StyleGAN2, StyleGAN3, Diffusion-GAN) the other is DM-based (ADM, DDPM, Score-SDE, GLIDE, Latent Diffusion, Stable Diffusion, DALL·E 2). These two sets comprise $2,500$ images per architecture, for a total of $17,500$ images each. Then we carry out an experiment in a closed set scenario (Fig.~\ref{fig:att_result}, top) and in an open set scenario (Fig.~\ref{fig:att_result}, bottom) where we limit the new unknown classes to come from 3 GAN-based generators (EG3D, GALIP, GigaGAN) and 3 DM-based generators for (DiT, DeepFloyd-IF, SDXL), respectively. In both scenarios, we consider $500$ images per generator under test. We can observe that in the closed-set scenario the performance is very good both for GANs and DMs. It worsens for open-set recognition, especially for DMs, where AUC can hardly reach 90\%. Interestingly, detection-based methods perform very well in this new setting, coherently with recent findings that the ability of a classifier to make a reject-option decision is very much correlated with its accuracy on the closed-set classes\cite{Vaze2022open}.

\subsubsection{Robustness analysis}
In the same scenario presented above, we also evaluate robustness to JPEG compression and blurring in order to check the sensitivity of the methods to possible benign perturbations.
Note that in this case, suitable augmentation is included in all the approaches as also suggested in the original papers. Results are presented in Fig.~\ref{fig:att_robust} and show that for most of the analyzed methods the decrease in performance is quite limited.

\subsubsection{Detection and attribution}
\begin{figure}[t!]
    \centering
    \setlength{\tabcolsep}{1pt}
    \begin{tabular}{lK{3.3cm}K{3.3cm}}
    \hspace{0.9cm} & Closed-set Accuracy  & Open-set AUC   \\
    \multicolumn{3}{l}{\includegraphics[width=0.95\linewidth,trim=0 0 0 0, clip,page=1]{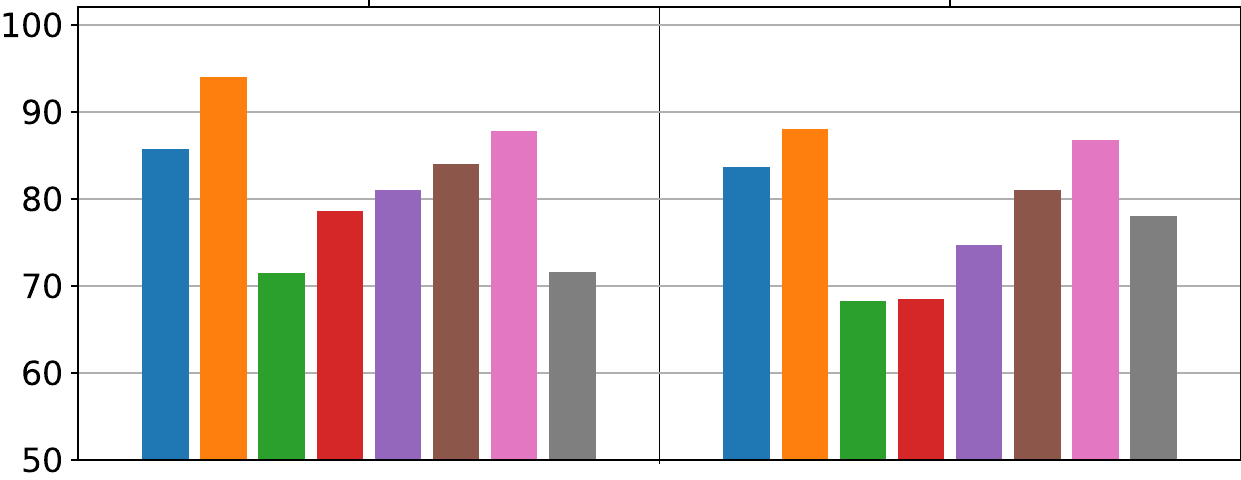}} \\
        & \multicolumn{2}{c}{\includegraphics[width=0.87\linewidth,trim=0 1 0 0, clip,page=2]{figures/results_att_big.pdf}}
    \end{tabular}
    \caption{Synthetic image attribution results in the more challenging scenario where also the real class is included.}
    \label{fig:att_wild}
\end{figure}

In this last experiment we consider a more challenging scenario where, beyond mixing GANs and DMs architectures, also real images are present. Note that in the literature, real images are separated based on their dataset of provenance. However, this may introduce biases that in turn help achieving a good performance and, eventually, lead to unfair comparisons. To be completely fair, we consider a single class that includes all real images, no matter what the original dataset.
In this situation, a very similar behavior is observed in the closed-set and open-set scenarios 
(Fig.~\ref{fig:att_wild}). All methods suffer only a small performance loss in the open-set scenario. Instead, a substantial loss is observed with respect to the situation where only synthetic images are considered (Fig.~\ref{fig:att_result}), especially in the closed-set scenario. It is also worth noting that the probability of correctly classifying the real class and the fake ones is balanced, with 94.8\% and 93.9\% respectively, for the best approach.

\section{Discussion and open challenges}
We are now in the condition to outline the main strengths of state-of-the-art methods for synthetic image verification as well as the current and future challenges that remain to be solved.
On the positive side we have:

\subsubsection{Low-level artifacts}
These forensic cues have been widely studied and exploited in several powerful methods. In the most favourable cases they allow for reliable detection and attribution. However, they provide valuable information also in more challenging situations. For example, they may help identifying the "family" of generative architectures used to generate the test image, thus restricting the search to a limited number of candidates. 

\subsubsection{Generalization}
In the favourable closed-set scenario, model attribution is extremely reliable. However, a very good performance is obtained also when the generators are used in slightly different conditions (different seed, loss function, training dataset). This means that if a malicious attacker relies on publicly available models, fine-tuning them on personal data, detection and attribution are still possible.

\subsubsection{Robustness}
There is plenty of experimental evidence that suitable forms of augmentation ensure good robustness to image impairments. Of course, it is important to know in advance the possible scenarios of interest. For example, if images are JPEG or Webp compressed, augmentation should be coherent with these formats, otherwise robustness is not necessarily guaranteed.

\vspace{2mm}
\setcounter{subsubsection}{0}
\noindent
There is also a number of problems yet to be solved:

\subsubsection{Detection vs. attribution} In the literature, these are often treated as separate and almost unrelated problems. Instead, they depend strongly on one-another, and should be dealt with jointly to optimize the performance.

\subsubsection{Open-set analysis}
Currently, in this scenario, sources that are not included in the training set are simply classified as unknown. However, many generative architectures are strongly correlated to one-another because they share common components. Therefore, there may be significant prior knowledge also on unknown samples which can be exploited to refine the analysis. Furthermore, it is worth highlighting that some methods suffer a large performance drop when moving from the closed-set to the open-set scenario.

\subsubsection{Calibration}
Results are often presented using average measures such as the AUC. However, a large AUC ensures only that two distributions can be well separated. The problem remains of how selecting the optimal decision threshold. The default threshold may provide dismaying results and hence some forms of calibration is needed to make decisions in a real-world scenario.

\section{Future directions}
As clear from the previous Section, there is still much room for research in this field, and there are many aspects that deserve deeper investigation.
Here, we outline only a few directions for future research, topics that, in our opinion, hold the most potential for real breakthroughs.   

\subsubsection{Intent characterization}
The boundary between real and fake is becoming thin. AI is already customarily used for compression, enhancement, super-resolution, and many more legitimate tasks. In the near future, generative AI will be everywhere. Should we keep calling these images "fake"?
In a world soon to be flooded by AI-generated content the major focus should be to characterize the intent of a media asset, be it real or generated: is it malicious or not? 

\subsubsection{Explainability}
Along a very similar path, the ability to explain the meaning of an image generated by artificial intelligence, especially in relation to its context, will allow to make sound decisions about its harmful potential. More in general, being able to provide an interpretation of the score provided by the detector would help to make more convincing decisions.
For example, it would be easier to trust a detector that can associate a sensible confidence level with its decisions.

\subsubsection{Robustness to adversarial attacks}
Although there are works that analyze the performance of detectors in the presence of adversarial attacks, only a few detectors are designed with the aim of withstanding such attacks. In particular, adversarial attacks can easily remove the low-level traces that many current detectors rely on.

\subsubsection{Universal approaches}
Beyond fully generated images, nowadays it is also possible to make local modifications to an image using a prompt, e.g. by adding/removing an object or even expanding the image. It would be desirable to design methods that can detect at the same time both global and local AI-generated content.

\subsubsection{Active methods}
In recent years, research has mainly focused on passive methods, neglecting active methods due to their well-known risks (e.g., privacy issues). However, modern active methods provide ingenious tools that should be considered to enrich the available forensic toolkit. Some approaches embed a watermark into an image in a visually imperceptible manner to certify image authenticity. Some other methods instead protect the images from malicious use and insert an invisibile signal with the purpose of disrupting editing tools and make them fail.

\vspace{2mm}
\noindent
It is difficult to predict whether these efforts will ensure the integrity of information in the era of generative artificial intelligence. However, this is an ongoing arms race, with neither side having a significant advantage. The availability of a wide variety of tools that follow different approaches and exploit complementary information is the main guarantee that ever new and reliable detectors can be designed and used to safeguard institutions and individuals.

\section{ACKNOWLEDGMENT}
We gratefully acknowledge the support of this research by a TUM-IAS Hans Fischer Senior Fellowship and a Google Gift. This material is also based on research sponsored by 
the Defense Advanced Research Projects Agency (DARPA) under agreement number FA8750-20-2-1004.
The U.S. Government is authorized to reproduce and distribute reprints for Governmental purposes notwithstanding any copyright notation thereon.
The views and conclusions contained herein are those of the authors and should not be interpreted as necessarily representing the official policies or endorsements, either expressed or implied, of DARPA or the U.S. Government.
In addition, this work has received funding by the European Union under the Horizon Europe vera.ai project, Grant Agreement number 101070093.
Finally, we want to thank Tero Karras, Yogesh Balaji, Ming-Yu Liu for sharing data for StyleGAN-T and eDiff-I experiments.

\balance
\bibliographystyle{IEEEbib-abbrev}
\bibliography{refs}

\end{document}